\documentclass[twocolumn,floatfix]{revtex4}
\usepackage[utf8]{inputenc}
\usepackage{mwe}
\usepackage{amsmath}
\usepackage{amsfonts}
\usepackage{algorithm}
\usepackage{algorithmicx}
\usepackage{algpseudocode}
\usepackage{hyperref}
\usepackage{cleveref}
\usepackage{subfig}

\normalsize
\usepackage{booktabs}

\DeclareMathOperator*{\argmax}{arg\,max\;}
\newcommand{\modeText}[1]{
\ifnum#1=0
    Network mode
\else\ifnum#1=3
    Flow mode
\else\ifnum#1=4
    Noodle mode
\fi\fi\fi
}

\begin{document}
\title{Mastering percolation-like games with deep learning}
\author{Michael M. Danziger}
\affiliation{AI for Healthcare, IBM Research, Haifa, Israel}

\author{Omkar R. Gojala}
\affiliation{Network Science Institute, Northeastern University, Boston, MA, 02115, United States}

\author{Sean P. Cornelius}
\affiliation{Department of Physics, Toronto Metropolitan University, Toronto, ON, M5B 2K3, Canada}
\affiliation{To whom correspondence should be addressed. \href{mailto:cornelius@toronotmu.ca}{cornelius@torontomu.ca}}

\begin{abstract}
Though robustness of networks to random attacks has been widely studied, 
intentional destruction by an intelligent agent is not tractable with previous methods.
Here we devise a single-player game on a lattice that mimics the logic of an attacker attempting to destroy a network.
The objective of the game is to disable all nodes in the fewest number of steps.
We develop a reinforcement learning approach using deep Q-learning that is capable of learning to play this game successfully, and in so doing, to optimally attack a network.
Because the learning algorithm is universal, we train agents on different definitions of robustness and compare the learned strategies.
We find that superficially similar definitions of robustness induce different strategies in the trained agent, implying that optimally attacking or defending a network is sensitive the particular objective.
Our method provides a new approach to understand network robustness, with potential applications to other discrete processes in disordered systems.
\end{abstract}
\maketitle

\section{Introduction}
In order for a networked system to be functional, the nodes need to be connected to one another. 
Though a small number of nodes may become disconnected without imperiling the system's overall performance, if the paths between nodes no longer exist on a macroscopic scale, the system will generally be non-functional. 

Percolation theory describes this transition from a connected network to a non-connected one having many small connected components. Early work on percolation treated the process of breaking connectivity as thermal, with sequential node or link failures occurring uniformly at random~\cite{albert-nature2000,cohen-prl2000}, thereby connecting the new theory of network science with older work in polymers, fractals and flow in inhomogeneous media~\cite{kirkpatrick-revmodphys1973,stauffer-1985,bundehavlin-1996}.
Later, Cohen et al. showed how percolation theory can be extended to non-random attack heuristics~\cite{cohen-prl2001}.
\begin{figure}
    \centering
    \includegraphics[width=\linewidth]{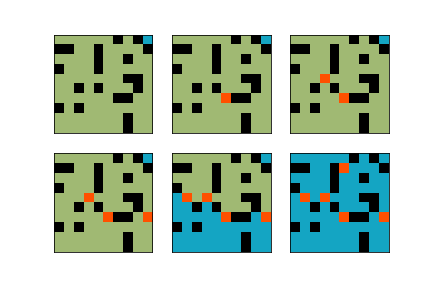}\\
    \includegraphics[width=0.8\linewidth]{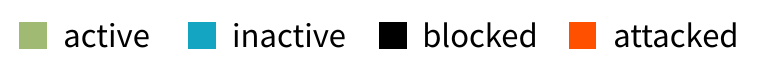}
    \caption{\textbf{A percolation-like game} 
    The aim of the game is to disconnect the green cells by ``attacking'' them (shown in red). If a component is disconnected from the largest connected component of green cells it is colored blue. When the largest remaining connected component of active (green) cells is smaller than one of the inactive (blue) components, it too becomes inactive and the game is over (lower right). For other percolation-like games, see Fig. \ref{fig:fig1}.}
    \label{fig:sample-game}
\end{figure}

\begin{figure*}
    \centering
    \includegraphics[width=\linewidth]{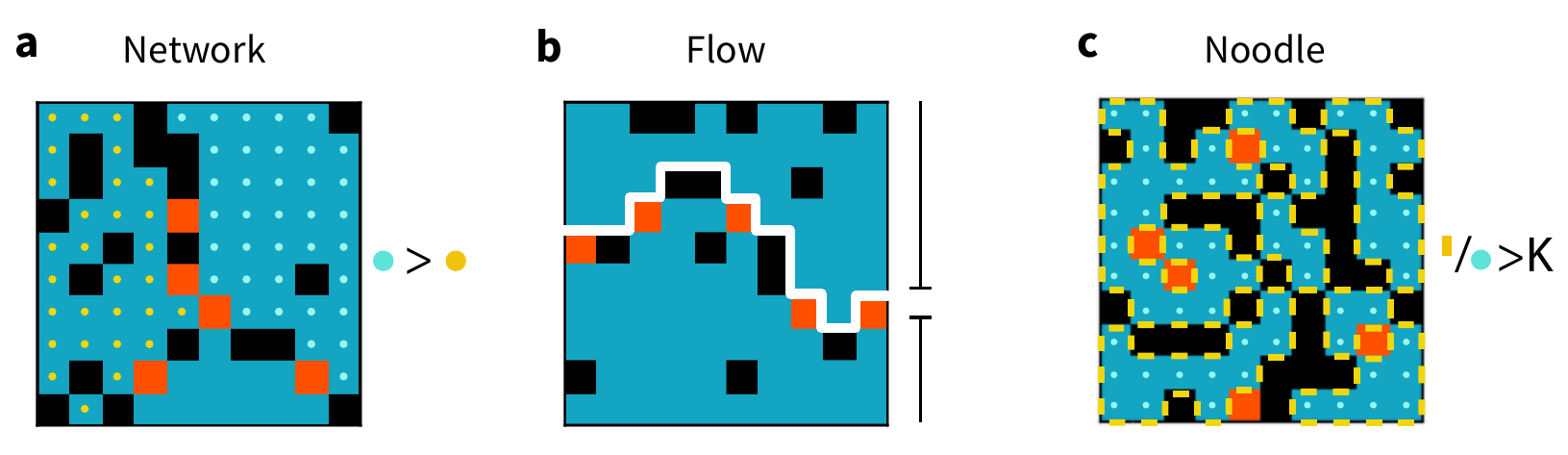}
    \caption{\textbf{Three percolation-like games.}
    \textbf{(a)} Network. The game ends when the largest component is made smaller than the second largest component. The dotted squares represent the board state in the step before the final move. The green dots represent the sites which had been active and the blue dots represent the second largest component.
    \textbf{(b)} Flow. The game ends when the passage from top to bottom is blocked.
    \textbf{(c)} Noodle. The game ends when the ratio of faces (tick marks) to (in)active squares, analogous to the ``surface area to volume'' ratio in this system, exceeds a specified ratio $K$.}
    \label{fig:fig1}
\end{figure*}

In recent years, there has been increasing interest not just in how a network responds to node removals under various heuristics, but rather attempting to discover the \emph{optimal} attack. 
Attempting to optimize the attack or defense of a network arises naturally in epidemiology~\cite{pastorsatorras-prl2001,cohen-prl2003,kim-lancet2015} and infrastructure resilience, respectively.
The closely related question of detecting influential nodes in an opinion spreading network has also been treated as a type of optimal percolation~\cite{kitsak-naturephysics2010,morone-nature2015,coghi-pre2018}.

Machine learning suggests a different approach for optimizing percolation strategy. Instead of defining analytically tractable heuristics, machine learning methods treat the problem as a ``black box,'' and use the expressive power of multilayer neural networks to optimize the objective. 
Deep reinforcement learning is particularly promising in this regard. 
The landmark AlphaGo~\cite{silver-nature2016} and AlphaZero~\cite{silver-nature2017} projects proved that deep reinforcement learning is capable of discovering superhuman strategies for Go, Chess and other classic board games. 
Go is of particular interest for our scenario, as winning the game requires one to create lattice connectivity more effectively than one's opponent, and can be analyzed in terms of percolation~\cite{yang-physicaa1991}.
Similarly, recent works in power grid resilience have used reinforcement learning to discover worst-case scenarios~\cite{yan-ieee2017,li-ieee2021,dehgani-appliedenergy2021}.

Here we formulate a simple single-player game over a square lattice, in which the objective is to disable all of the nodes in as few steps as possible.
The squares of the board represent nodes of a network, each being connected to its four nearest neighbors.
Each square can be in one of four states: active (green), inactive (blue), attacked (red) and blocked (black).
At the onset, the non-black squares form one or more components based on their network connectivity.
At each step, the player can disable an active node, turning it from green to red, as shown in Fig. \ref{fig:sample-game}. 
The red squares are functionally equivalent to black squares.
This affects the overall connectivity of the system, and may disconnect a component, turning all its squares blue.
The game ends when all of the components have turned blue, \emph{i.e.} lost connectivity. 

When defining a percolation-like game, the simplest criterion of membership in the largest connected component is not useful, as some component will always be largest, and the game will end only when all nodes have been eliminated.
To make a percolation-like game we begin by introducing two distinct game modes, corresponding to different ending criteria: network mode and flow mode.

In network mode (Fig. \ref{fig:fig1}a) the nodes are active as long as they are in the largest connected component, \emph{and} that component is larger than the largest of the inactive components.
This is comparable to the transition in which the second-largest component overtakes the largest component in standard percolation, which has been utilized for difficult to measure percolation transitions~\cite{li-pnas2015}, and divergence of the second-component size is observed at criticality~\cite{margolina-physlettersa1982}.
With this objective, the player can win more quickly by keeping the largest inactive component large, and not try to cause maximal disconnection at each step.

In flow mode (Fig. \ref{fig:fig1}b), we assume a lattice structure and the condition for remaining active is that the node be a member of a connected component with nodes on top and bottom of the lattice, as if the nodes were resistors and the edges of the lattice were fixed voltage sources. 
While this mode assumes a specific physical structure, network mode is well-defined on any network.
We use lattices here for simplicity and comparability across modes.

\begin{figure}
    \includegraphics[width=\linewidth]{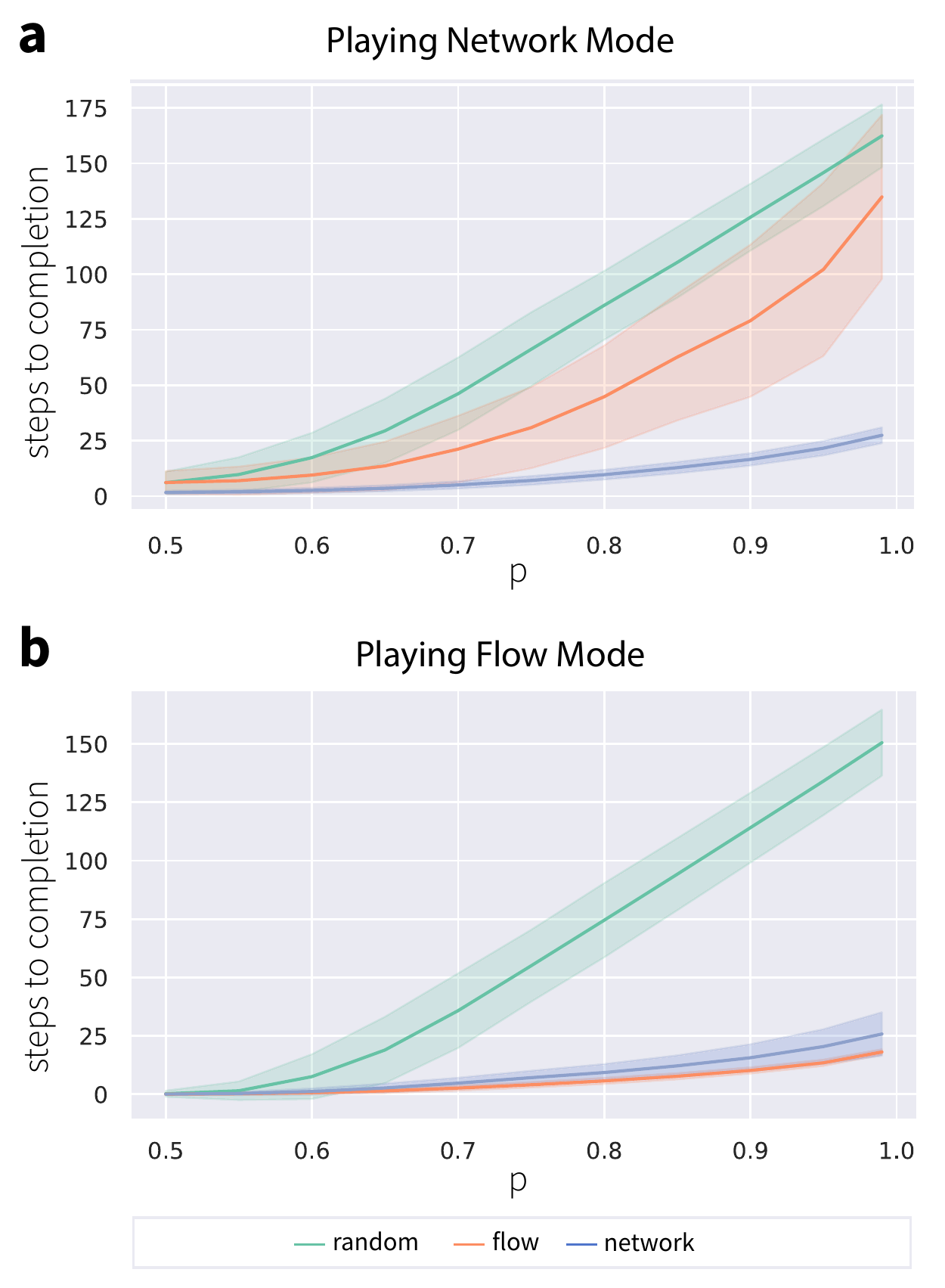}%
    \caption{\textbf{Number of steps to completion for agents trained on the flow task and the network task, compared to random moves, playing the network mode \emph{(a)} and the flow mode \emph{(b)}}.
    As the boards become more open (higher $p$), more moves are required to complete the game. 
    The random curve represents the standard percolation process.
    In panel \textbf{(a)}, we see that the flow-trained agent performs poorly on the network task, while the network-trained agent performs well on both tasks, see Fig. \ref{fig:modevmode} for further details.
    Solid lines denote the average performance over games on 1,000 randomly-generated 20 $\times$ 20 boards with the given probability $p$, with standard deviation for error bars.}
    \label{fig:playing_network_flow}
\end{figure}

\begin{figure}
    \includegraphics[width=\linewidth]{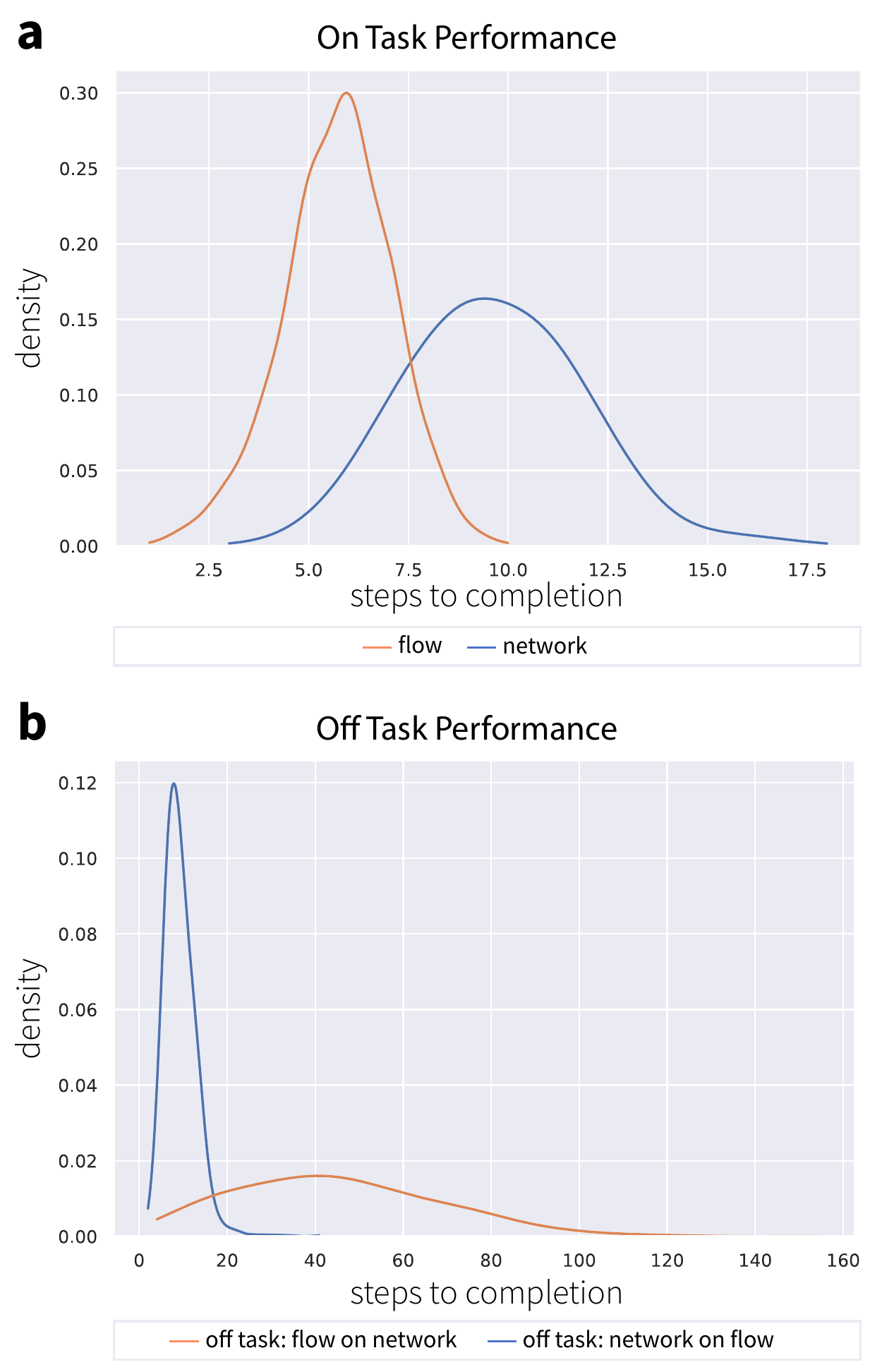}%
    \caption{\textbf{Comparison of performance for different agents.}
    While the agent trained on the flow objective completes the flow objective faster, it generalizes poorly when it attempts the network objective. 
    In contrast, the agent trained on the network objective takes longer to complete the network objective but generalizes better to the flow objective.}
    \label{fig:modevmode}
\end{figure}
The objective of the game is to make all the nodes disconnected (shown in blue) in as few moves as possible, see Fig. \ref{fig:sample-game} for an example game. 
In contrast to previous work on network dismantling which seeks to optimize cumulative efficiency of the dismantling process~\cite{braunstein-pnas2016,ren-pnas2019,fan-nmachineintel2020,grassia-ncomms2021}, we focus on objectives that minimize the steps to completion. 
Our objectives are thus more similar to to golf than the two-player competition found in other percolation-like games such as Hex~\cite{hefetz-2014} or Go.

We have tested the game extensively with human players of all ages, who found its solution to be learnable, though not trivial. The game can be played freely at \url{http://blue.mmdanziger.com/}.

\section{Results}
Given a well-defined objective such as the percolation-like games defined above, we train an agent to master the game using deep convolutional networks and Q-learning, a form of reinforcement learning. The goal here is to learn a \emph{Q-function}:
\begin{equation}
Q(s, a),
\label{eq:q-function}
\end{equation}
which encodes the total future reward that can be accrued by playing an action $a$ in game state $s$, assuming optimal subsequent play. In our setting, $s$ is the current game board, and the eligible actions $a$  are all squares in $s$ whose state is ``alive'' (green, in our color scheme). In possession of such a function, the agent plays according to a greedy strategy, always choosing the best possible action $a^*$ in the given game state, namely
\begin{equation}
a^* = \argmax_a Q(s, a).
\label{eq:q-policy}
\end{equation}

We parameterize the (unknown) $Q$-function using a deep convolutional neural network, which we train for a given game mode via self-play on randomly-initialized game boards of size 20 $\times$ 20, with a range of initial densities (vis-\`{a}-vis the parameter $p$). For details of the network architecture and training algorithm, we refer the reader to \emph{Methods}.

\begin{figure}
    \centering
    \includegraphics[width=\linewidth]{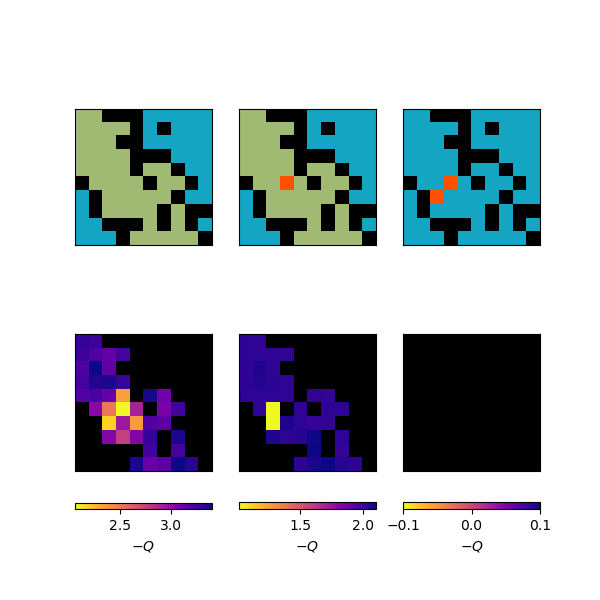}
    \caption{\textbf{Illustration of gameplay in terms of Q-values}
    For each stage of the game (top row), the trained agent calculates a Q value, corresponding to the move that minimizes the objective, as shown in the bottom row.
    While for the first move, the agent identifies several potentially good moves, the second move has exactly two moves, either of which will end the game. 
    Example shown for agent playing the network mode with initial open spaces of 0.75. 
    See Supplementary Figures S1-S15 for more examples.
    }
    \label{fig:q_learning}
\end{figure}

After training a deep-learning agent for each game mode, we first analyze their on-task performance. We find that trained agents significantly outperform random move selection for both network (Fig. \ref{fig:playing_network_flow}a) and flow (Fig. \ref{fig:playing_network_flow}b) modes.
The difference between the trained agents and random play is not substantial when the initial board is close to the percolation threshold ($p \approx 0.6$), but their performance diverges as the sparseness increases ($p \rightarrow 1$).
This is expected, as near criticality, the likelihood of a random move ending the game becomes very high.

Given that $Q$-learning produces a $Q$-function, which estimates the best-case total moves to completion for each candidate action, we can gain insight into the agent's strategy by visualizing the $Q$ value of each square at different stages of a game. An example for ``network'' mode is shown in Fig. \ref{fig:q_learning}, and more examples are presented in Supplementary Figures S1-S15.
We see that the agent frequently identifies multiple moves of essentially equal value.
We also find that it correctly learns to maximize the largest inactive component size as the most effective path to success in this mode. In particular the agent correctly identifies ``choke points'' bridging large active (green) components as high-value moves, whose play will win the game quickly.

Having agents trained on different objectives, it is instructive to examine their performance when attempting the task they were not trained for.

We find that the machine strategy for network mode is also performant in flow mode, though somewhat less effective (\ref{fig:playing_network_flow}b and Fig. \ref{fig:modevmode}a).
But interestingly, the converse is not true (\ref{fig:playing_network_flow}a).
Though the flow agent often solves the network game in a reasonable number of steps, it can also get stuck and require a very large number of steps to completion (Fig. \ref{fig:modevmode}b).
This is understandable because the optimal strategy for flow mode is simply to create any horizontal ``barrier'' across the lattice, but for network mode this strategy\textemdash while it works in some cases\textemdash is not a valid general solution.
The fact that the network strategy proves good enough for the flow mode, while the flow strategy fails for the network mode suggests a hierarchy of difficulty.

\begin{figure}
    \centering
    \includegraphics[width=0.85\linewidth]{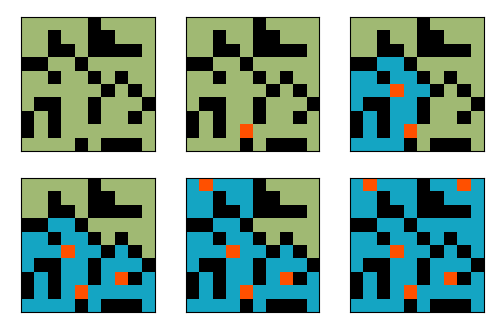}\\
    \includegraphics[width=0.85\linewidth]{actualLegend.pdf}
    \caption{\textbf{Noodle mode gameplay.} The objective here is to bring the ratio of surface area (number of faces shared with boundaries) to volume (number of squares) for each component below $K=2$. This objective is less intuitive, yet still learnable, as shown in Fig. \ref{fig:noodleofftask}.}
    \label{fig:noodle}
\end{figure}

\begin{figure}
        \includegraphics[width=\linewidth]{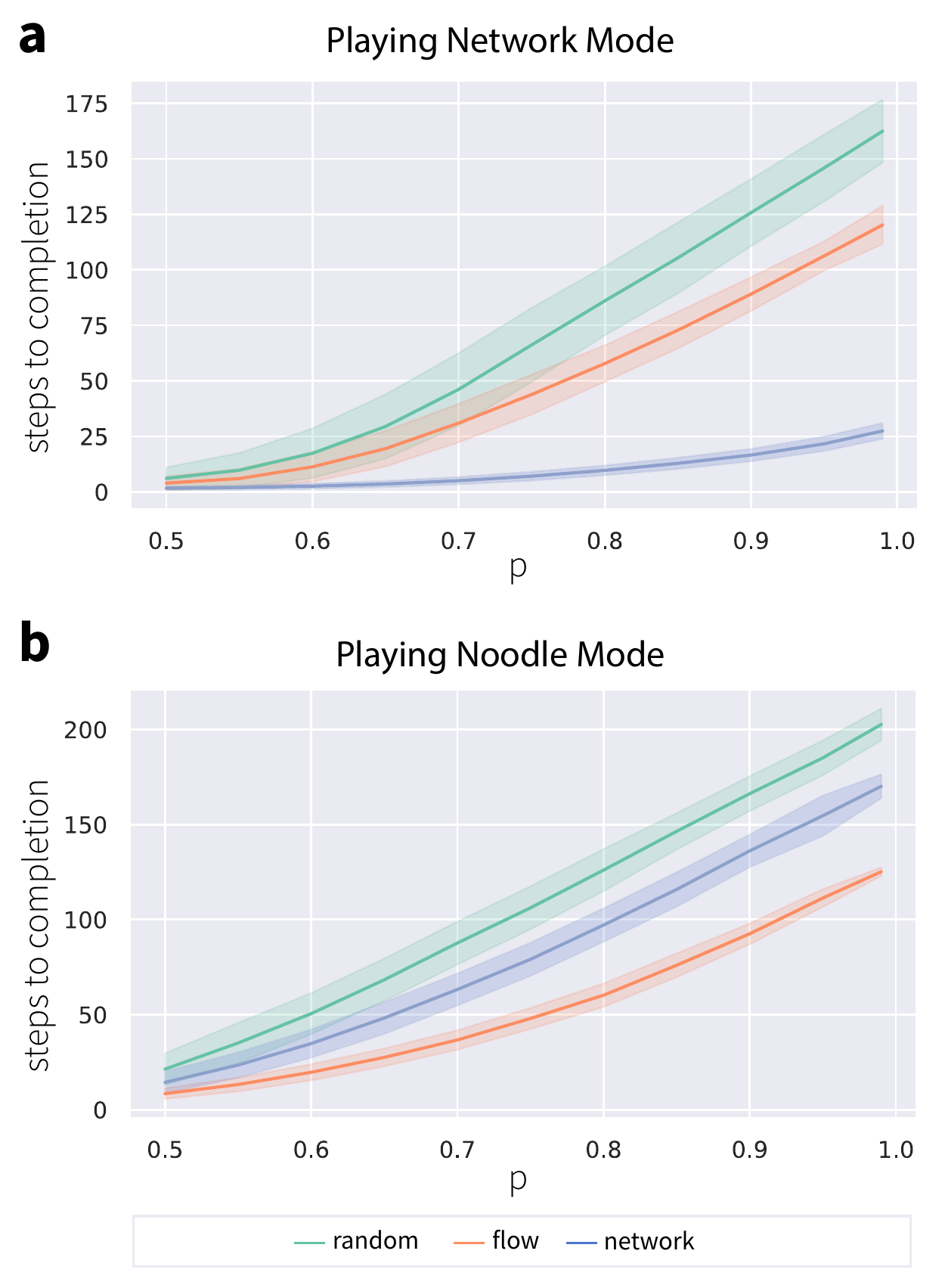}%
\caption{\textbf{Number of steps to completion for agents trained on the noodle task and the network task, compared to random moves, playing the network mode \emph{(a)} and the noodle mode \emph{(b)}.}
    In contrast to Fig. \ref{fig:modevmode}, where we showed the network-trained agent performing well on both network and flow modes, here we find that neither the network nor the noodle agent performs well on the others' task.
    Results are obtained from 1000 simulated games on a 20x20 board, with standard deviation for error bars.}
    \label{fig:noodleofftask}
\end{figure}
Until now, we have considered two game modes in which connectivity is defined in ways similar to conventional percolation.
However, our learning framework can explore a far more diverse set of objectives.
As an example, we consider a new objective inspired by surface energy in which the a component become inactive if the ``surface area to volume'' ratio of their component exceeds a given threshold $K$.
Specifically, we compute the ratio of the number of faces on the boundary of a component and the sites contained within, as in shown Fig. \ref{fig:fig1}c. The component is marked inactive if this ratio is greater than $K$.
The surface area of percolation components has been studied under random attacks as cluster hulls~\cite{leath-jphysicsc1978}, perimeters~\cite{voss-jphysicsa1984} and the surface fractal dimension~\cite{vanderzande-jphysicsa1988}.
Because the agent must increase the circuitousness within the components, we call this objective ``noodle'' mode. 

Even though this reflects a substantially different objective compared to the other modes and is not even based on connectivity, our deep learning architecture can train an agent to master this game, as we show in Figure \ref{fig:noodle}.
Interestingly, we find that neither the noodle agent nor the network agent is particularly successful at the other's task (Fig. \ref{fig:noodleofftask}).
The training process correctly identifies these tasks as non-commensurate, (in contrast to flow and network) suggesting the discoverability of a non-trivial hierarchy of task difficulty.

\section{Discussion}

Network resilience under attack is a problem that arises in multiple contexts, and yet conventional analysis treats the problem without allowing for the intentionality of the attacking agent.
Here we have shown how the problem of network robustness can be re-cast as a playable game, enabling us to explore the behavior of human or machine agents.
The game itself has been enjoyed by many people and may prove valuable as a tool for science education.
In particular, defining the end state as the moment that the second largest component becomes larger than the largest component allows us to conserve the basic logic of percolation while delivering a non-trivial game, which can be accessed at \url{http://blue.mmdanziger.com/}.
The fact that the game constitutes a fresh yet realistic task also enables future research comparing human and machine patterns of learning.

This gamification enables us to define clear Q-learning objectives for percolation, and to train an agent to win the game with deep reinforcement learning. 
Though we began with standard percolation, by utilizing Q-learning we are able to generalize to unusual objectives like ``noodle'' mode. 
As such, we anticipate further interest in optimizing discrete time processes which are not amenable to conventional differential equation representations.

For the purposes of simplicity, we have limited ourselves to lattice topologies in this work. However, our approach can also be generalized to more complex topologies using the same Q-learning framework by making use of graph convolutional networks~\cite{kipf-arxiv2016}.

And while we have focused on the attack scenario, similar objectives can be defined for a defensive scenario, where the objective is to maintain or restore connectivity or functionality~\cite{sanhedrai-nphys2022}.
This enables a fusion of network resilience and deep learning to model the complex adaptive games of attack and defense in natural and human systems.

\clearpage

\bibliography{blue.bib}
 
\clearpage
\section{Materials and Methods}
\label{sec:methods}

\subsection{Deep Q-Learning}
\label{sec:deepq}

\par \noindent \textbf{The Bellman Equation}. By its definition, the $Q$-function must obey a self-consistent recurrence relation (the Bellman equation), namely
\begin{equation}
Q(s, a) = r(s, a) + \max_{a'} Q(s', a').
\label{eq:bellman}
\end{equation}
Here, $s'$ is the new game state that would result from playing the action in question ($a$) in the current state ($s$), and $r(s, a)$ is the immediate (incremental) reward obtained in the process. In our setting, we use $r(s, a) = -1$ in all game modes, meaning the agent receives a penalty for every move played. By playing to maximize $Q$ (cf.~\cref{eq:q-policy}), the agent thus strives to complete the game in the minimum number of moves. 
By definition, we have the boundary condition
\begin{equation}
   Q(s^*, a) = 0 
\label{eq:boundary-condition}
\end{equation}
for any terminal game state $s^*$ (when all connected components are blue, in our color scheme). Together with \cref{eq:bellman}, this allows us to model $Q(s, a)$ by fitting an appropriately general ``universal approximator'' function. \\

\par \noindent \textbf{Network Architecture}. We parameterize $Q(s, a)$ using a Deep-Q network (DQN), consisting of two stages. First, a \emph{feature embedding}, which takes as input a $n \times n \times 4$ boolean tensor, corresponding to a one-hot encoding of the status of each of the $n \times n$ squares. It outputs a $n \times n \times m$ tensor $x$, encoding the state of each of the $n^2$ cells $(i, j)$ as an $m$-dimensional \emph{feature vector}, $x_{ij}$. The embedding is performed by a $d$-layer convolutional neural network, where each layer consists of a $3 \times 3$ convolution followed by a ReLU nonlinearity. Here $d$ and $m$ are hyperparameters.

Given the embedded board state $x$, we evaluate $Q(s, a)$ by a function $\hat{Q}(v, x_{ij})$, which uses $x_{ij}$ as a proxy for the action $a = (i, j)$, and a vector of \emph{global} (pooled) features $v$ as a proxy for the board state as a whole ($s$). Specifically, we take $v$ to be
a $4m$ length vector comprised of the minimum, maximum, sum, and average of $x$, where each operation is applied channel-wise to each of the $m$ features in $x$. We then calculate the $Q$ value according to:
$$
Q(s, a) = \hat{Q}(v, x_{ij}) = w_1^T \textnormal{relu} \left[w_2 v, w_3 
 x_{ij} \right],
$$
where we set $Q(s, a) = -\infty$ for ineligible actions (non-green squares) $a$, ensuring they won't be chosen under the greedy policy defined by \cref{eq:q-policy}.
Here, $w_1 \in \mathbb{R}^{5m}$, $w_2 \in \mathbb{R}^{4m \times 4m}$, and $w_2 \in \mathbb{R}^{m \times m}$ 
are trainable weights.  As such, $Q(s, a)$ depends on a large number of parameters: $w_1$, $w_2$, and $w_3$ as well as the parameters of each convolutional layer in the embedding. The values of these parameters that produce the best fit to Eqs.~\ref{eq:bellman}-\ref{eq:boundary-condition} are learned via self-play.
 \\
\begin{table}[th]
\begin{tabular}{ccc}
\toprule
\textbf{Hyperparameter} & \textbf{Meaning} & \textbf{Value} \\ \midrule
$d$ & depth & 10 \\
$m$ & embedding dimension & 64 \\
$N_\textnormal{replay}$ & replay capacity & $10^6$ \\
$T_\textnormal{rollout}$ & rollout frequency & $10^3$ \\
$N_\textnormal{rollout}$ & games per rollout & $10$ \\
$\epsilon_{\max}$ & initial exploration probability & $1.00$ \\
$\epsilon_{\min}$ & final exploration probability & $0.05$ \\
$T_\textnormal{anneal}$ & $\epsilon$ annealing period & $10^5$ \\
$N_\textnormal{batch}$ & batch size & 128 \\
$T_{\max}$ & total train epochs & $10^6$ \\
$T_\text{update}$ & target network update frequency & $10^3$ \\
\bottomrule
\end{tabular}
\caption{\textbf{Hyperparameters.}}.
\label{table:hyper}
\end{table}
\par \noindent \textbf{Training}. We train each agent on synthetic data (see \cref{sec:methods}---Data Generation) as described in \Cref{alg:algorithm}. At every step (epoch) of the training process, we update the DQN weights via propagation using the Adam optimizer to minimize the mean-squared error between the left- and right-hand sides of \cref{eq:bellman}. We take advantage of two techniques to improve training: \emph{experience replay} and \emph{double Q learning}.

In \emph{experience replay}, we keep a buffer $\mathcal{M}$ containing the most recent $N_\text{replay}$ experiences $(s, a, s', r)$, where $N_\text{replay}$ is a hyperparameter. At every training epoch, we sample a batch of $N_\text{batch}$ experiences uniformly from the buffer, and evalute the left- and right hand sides of \cref{eq:bellman} over this batch, and update the weights accordingly. As the network is trained, we periodically add new experiences to the replay buffer by ``rolling out'' the updated policy. Specifically, every $T_\text{rollout}$ epochs, we play $N_\text{rollout}$ complete games from randomly-initialized boards, and add those experiences to the replay buffer. To balance exploration and exploitation, we use an $\epsilon$-greedy policy for action selection. At every step of a game, the agent agent selects a random eligible action with probability $\epsilon$, or greedily according to the current $Q$-function (via \cref{eq:q-policy}) with probability $1 - \epsilon$. The exploration probability $\epsilon$ is decreased linearly from $\epsilon_\text{max}$ to $\epsilon_{min}$ over the first $T_\text{anneal}$ epochs of training, remaining at $\epsilon_{min}$ thereafter. Here $\epsilon_\text{max} > \epsilon_{min}$ are hyperparameters. 

\emph{Double Q learning} helps tackle the overestimation of Q values in traditional Q learning \cite{hasselt2016deep}, it uses an intermediary technique called fixed Q targets. In this approach, one keeps two copies of the DQN: a \emph{policy} network, and a \emph{target} network. The policy network is used to play the game (select actions) and hence evaluate the left hand side of \cref{eq:bellman}, $Q(s, a)$, and has its weights updated every iteration of training. In contrast, the target network is kept largely static, being used only to evaluate the right hand side of \cref{eq:bellman}. Every $T_\text{update}$ iterations, we update the target network by copying the weights from the policy network, where $T_\text{update} \gg 1$ is a hyperparameter. 

The values of all hyperparameters used in this study are listed in \Cref{table:hyper}. \\


\begin{algorithm*}
\caption{Training}
\begin{algorithmic}
\State Initialize the replay memory $\mathcal{M}$ to capacity $N_\textnormal{replay}$
\For{Epoch $e = 1$ to $L$}
\If{$e \bmod T_\textnormal{rollout} = 0$}
\State Generate a set $S$ of $N_\textnormal{rollout}$ random boards
\For{Board $s \in S$}
\While{$s$ is not terminal}
\State{Select an action $a$ according to
$$
a = \begin{cases}
\textnormal{Random eligible square} & \textnormal{w.p. } \epsilon \\
\displaystyle \argmax_a Q(s, a) & \textnormal{w.p. } 1-\epsilon
\end{cases}
$$
}
\State Play $a$, yielding incremental reward $r$ and next state $s'$
\State Add experience $(s, a, s', r)$ to $\mathcal{M}$
\State{$s \gets s'$}
\EndWhile
\EndFor
\EndIf
\State Sample a batch of experiences $B$ of size $N_\textnormal{batch}$ uniformly from $\mathcal{M}$
\State Update the weights $\mathbf{w}$ over $B$ via gradient descent on \cref{eq:bellman}
\EndFor
\end{algorithmic}
\label{alg:algorithm}
\end{algorithm*}

\noindent \textbf{Software implementation}. All simulations in this study were performed in
Python, using the \texttt{pytorch} library for deep learning. Our source code is freely available online at \url{https://github.com/spcornelius/blue_zero}. \\

\noindent \textbf{Data generation}. We generate training/validation data by randomly initializing board states of sizes 20 $\times$ 20 ($n = 20$) with a fraction $1 - p$ of ``blocked'' (black) squares. The remaining squares are set to ``active'' (green) or ``inactive'' (blue) according to the rules of the given game mode. We discard any boards that represent terminal states. We sample $p$ uniformly between [0.5, 1.0]. As such, the agent for each game mode is trained and validated boards over a range of fill densities.

\section{Acknowledgments}
S.P.C. acknowledges support from the Natural Sciences and Engineering Research Council of Canada (NSERC, grant no. RGPIN-2020-05015). The authors thank O. Varol, A. Grishchenko, and X. Meng for helpful discussions.

\end{document}


\title{Supplementary Information for: \\
Mastering percolation-like games with deep learning}
\maketitle

\renewcommand{\thefigure}{S\arabic{figure}}
\renewcommand{\thesection}{S\arabic{section}}
\setcounter{figure}{0}    

\section{Example games}
In this section we bring several examples of gameplay across network, flow and noodle modes.

\fig{0}{10}{0.7}{0}
\fig{0}{10}{0.7}{1}
\fig{0}{10}{0.7}{2}
\fig{0}{10}{0.7}{3}
\fig{0}{10}{0.7}{4}

\fig{3}{10}{0.8}{10}
\fig{3}{10}{0.8}{11}
\fig{3}{10}{0.8}{12}
\fig{3}{10}{0.8}{13}
\fig{3}{10}{0.8}{14}

\fig{4}{10}{0.6}{30}
\fig{4}{10}{0.6}{31}
\fig{4}{10}{0.6}{32}
\fig{4}{10}{0.6}{33}
\fig{4}{10}{0.6}{34}